\pdfoutput=1
\documentclass{article}
\usepackage[T1]{fontenc}
\usepackage{times}
\usepackage[left=34mm,right=34mm,top=33mm,bottom=20mm]{geometry}
\usepackage{natbib}
\usepackage{booktabs}
\usepackage{myauthor}

\usepackage{url}
\usepackage{amsmath,amssymb,amsfonts}
\usepackage[amsmath,thmmarks,amsthm]{ntheorem}
\usepackage{bbm} 

\usepackage[ruled,vlined,longend]{algorithm2e}

\usepackage{floatrow}
\newfloatcommand{capbtabbox}{table}[][\FBwidth]
\usepackage{blindtext}

\usepackage{enumitem}
\usepackage{graphicx}
\usepackage{verbatim}
\usepackage{bm}
\usepackage{mathrsfs}
\usepackage{aliascnt}
\usepackage{xcolor}
\usepackage[colorlinks,
            linkcolor=blue,
            citecolor=blue,
            urlcolor=magenta,
            linktocpage,
            plainpages=false]{hyperref}
\usepackage{url}
\usepackage[capitalize,nameinlink]{cleveref}

\bibpunct{(}{)}{;}{a}{,}{,}

\DeclareMathOperator*{\argmin}{arg\,min}

\DeclareMathOperator*{\tr}{\mathrm{tr}}

\newcommand{\real}{\mathbb{R}}

\newcommand{\X}{\mathcal{X}}

\newcommand{\diag}{\mathrm{diag}}

\definecolor{dark-red}{rgb}{0.9,0,0}
\definecolor{dark-green}{rgb}{0,0.6,0}
\definecolor{dark-blue}{rgb}{0,0,0.85}

\newcommand{\image}{\mathrm{Im}}
\newcommand{\Pperp}{P^\perp}
\newcommand{\tPi}{\tilde{\Pi}}

\newcommand{\regret}{\mathcal{R}}

\newtheorem{conjecture}{Conjecture}

\definecolor{light-gray}{gray}{0.9}

\qedsymbol{\ensuremath{_\blacksquare}}

\newaliascnt{theorem}{thm}
\newtheorem{theorem}[theorem]{Theorem}
\aliascntresetthe{theorem}

\newaliascnt{corollary}{thm}

\aliascntresetthe{corollary}

\newaliascnt{lemma}{thm}
\newtheorem{lemma}[lemma]{Lemma}
\aliascntresetthe{lemma}

\newaliascnt{example}{thm}

\aliascntresetthe{example}

\newaliascnt{proposition}{thm}

\aliascntresetthe{proposition}

\newaliascnt{definition}{thm}

\aliascntresetthe{definition}

\newcommand{\BlackBox}{\rule{1.5ex}{1.5ex}}  
\newenvironment{proof-sketch}{\par\noindent{\bfseries\upshape
  Proof\ sketch\ }}{\hfill\BlackBox\\[2mm]}

\title{CompAdaGrad: A Compressed, Complementary, Computationally-Efficient Adaptive Gradient Method}

\author{
Nishant A.~Mehta\thanks{Research conducted while at Australian National University} \\
Centrum Wiskunde \& Informatica \\
\href{mailto:mehta@cwi.nl}{mehta@cwi.nl} \\
\And
Alistair Rendell \\
Australian National University \\
\href{mailto:Alistair.Rendell@anu.edu.au}{Alistair.Rendell@anu.edu.au} \\
\AND
Anish Varghese \\
Australian National University \\
\href{mailto:anish.varghese@anu.edu.au}{anish.varghese@anu.edu.au} \\
\And
Christfried Webers\thanks{Research conducted while at NICTA \& Australian National University} \\
Data61 \& Australian National University \\
\href{mailto:christfried.webers@data61.csiro.au}{christfried.webers@data61.csiro.au} \\
}

\begin{document}

\maketitle

\begin{abstract}
The adaptive gradient online learning method known as AdaGrad has seen widespread use in the machine learning community in stochastic and adversarial online learning problems and more recently in deep learning methods. The method's full-matrix incarnation offers much better theoretical guarantees and potentially better empirical performance than its diagonal version; however, this version is computationally prohibitive and so the simpler diagonal version often is used in practice. We introduce a new method, CompAdaGrad, that navigates the space between these two schemes and show that this method can yield results much better than diagonal AdaGrad while avoiding the (effectively intractable) $O(n^3)$ computational complexity of full-matrix AdaGrad for dimension $n$. CompAdaGrad essentially performs full-matrix regularization in a low-dimensional subspace while performing diagonal regularization in the complementary subspace. We derive CompAdaGrad's updates for composite mirror descent in case of the squared $\ell_2$ norm and the $\ell_1$ norm, demonstrate that its complexity per iteration is linear in the dimension, and establish guarantees for the method independent of the choice of composite regularizer. Finally, we show preliminary results on several datasets.
\end{abstract}

\section{Introduction}
\label{sec:intro}

Modern machine learning applications often involve high-dimensional datasets with large sample sizes, on which simple algorithms such as variants of online gradient descent are competitive with more complicated batch algorithms. 
In addition to often being more practical computationally as compared to batch methods, online methods apply to a wider range of scenarios such as online prediction against individual sequences. 
In the online non-stochastic (adversarial) setting, the regret of a learning algorithm is a more natural quantity to analyze than the cumulative loss (for the latter an adversary may as well emit a data sequence of pure noise). 
Although vanilla online gradient descent \citep{zinkevich2003online} obtains provably optimal regret in a minimax sense \citep{abernethy2008optimal}, if a method can simultaneously admit better guarantees against easy data sequences (such as low-dimensional data or i.i.d.~stochastic data) while maintaining the fallback guarantee of minimax optimality against the nastiest data sequences, such a method is even better. Such methods fall within the recently-sculpted space of ``learning faster from easy data'' \citep{grunwald2013learning}.

AdaGrad \citep{duchi2011adaptive} embodies an adaptive gradient family of algorithms, the two most prominent cases being full-matrix AdaGrad and diagonal AdaGrad. Whereas the former is computationally intractable in high-dimensions, the latter is tractable with only linear complexity in the dimension of the data. Both methods admit sequence-dependent regret bounds that can be much better than the regret bounds of online gradient descent. However, the computational levity of the diagonal version is matched with a price: if the components of the gradient are highly correlated, the diagonal version may fail to adapt well while the full-matrix version continues to be adaptive.

In addition to its theoretical strengths, even the diagonal version of AdaGrad has exhibited strong empirical performance on real-world problems \citep{duchi2011adaptive}. Moreover, diagonal AdaGrad also has been incorporated into deep learning algorithms that currently achieve state-of-the-art learning performance on a number of difficult image classification tasks \citep{dean2012large}.

Even though it is theoretically superior in terms of the regret, the full-matrix version has not seen similar large-scale applications due to its high per-round complexity of $O(n^3)$. This raises the question: is it possible to design a method lying between the full-matrix and diagonal versions of AdaGrad whose per-round complexity is $O(n)$? In this work, we answer this question in the affirmative, up to log factors, by presenting CompAdaGrad. 
This method replaces full-matrix AdaGrad's Bregman divergence regularization with the sum of a \emph{compressed} Bregman divergence operating in a low-dimensional subspace and a diagonal Bregman divergence operating in the orthogonal complement of that subspace. The precise form can be seen in \eqref{eqn:psi-rp-oc}. This method admits a theoretical guarantee that appears to be between full-matrix AdaGrad and diagonal AdaGrad, as shown in \cref{sec:regret}.

It is natural to ask why one would compress at the regularization level rather than simply compressing the data itself via a random projection from the very beginning. There are several reasons why compression at the regularization level makes more sense.
\begin{enumerate}
\item An initial compression of the data can reduce computational complexity but may lose information irretrievably, whereas by shifting the compression into the regularization component of the objective some part of the otherwise lost complementary information can still be exploited.
\item In an adversarial setting, if the learning algorithm commits to a fixed low-dimensional subspace, even an oblivious adversary can ensure that all the interesting action occurs in the complement of this subspace.
\item In some applications, one actually needs a predictor in the original space. This might be for interpretability, for instance.
\end{enumerate}

In the next section, we review AdaGrad and derive CompAdaGrad. In \cref{sec:regret} we present a regret bound for CompAdaGrad. We show how to compute updates for certain composite regularizers in \cref{sec:computations}. This section also contains a result that may be of independent interest: \cref{thm:1-sparse} establishes $O(n)$ complexity for computing the $n$-dimensional Walsh-Hadamard Transform of a 1-sparse vector. 
In \cref{sec:experiments}, we present experimental results on several datasets. Finally, we conclude the paper.

\section{AdaGrad: The Full, the Diagonal, and the Compressed}
\label{sec:adagrad}

\paragraph{Notation.}
Throughout this paper, we let $\X = \real^n$ and leave extensions to convex subsets for future work. 
For a strongly convex, differentiable function $\psi$, let $B_\psi$ be the Bregman divergence induced from $\psi$, defined as
\begin{align*}
B_\psi(x, y) = \psi(x) - \psi(y) - \langle \nabla \psi(y), x - y \rangle .
\end{align*}

For a sequence of subdifferentiable convex functions $f_1, \ldots, f_T$, let $g_1, \ldots, g_T$ be a corresponding sequence of subgradients, so that for each $t \in [T]$ we have $g_t \in \partial f_t$.

\paragraph{Online Convex Optimization game.}

We consider the following online learning protocol parameterized by a convex regularization function $\varphi: \X \rightarrow \real_+$.

Let Nature be an oblivious adversary; that is, before the game begins Nature selects its sequence of functions possibly with knowledge of Learner's (potentially randomized) strategy. The game then proceeds over a sequence of rounds:

\begin{quote}
For round $t = 1, 2, ...$
\begin{enumerate}[label=(\arabic*)]
\item Learner makes a prediction $x_t$ in action space $\X$.
\item Nature reveals a convex loss function $f_t : \X \rightarrow \real_+$ .
\item Learner suffers composite loss $f_t(x_t) + \varphi(x_t)$.
\end{enumerate}
\end{quote}

The goal is to find a strategy that minimizes the regret, defined as follows. 
Let $f_1, \ldots, f_T$ be a sequence of functions chosen by an oblivious adversary.
Then the regret on this sequence is
\begin{align} \label{eqn:regret}
\regret(f_1, \ldots, f_T) 
= \sum_{t=1}^T \bigl( f_t(x_t) + \varphi(x_t) \bigr) 
    - \inf_{x^* \in \X} \sum_{t=1}^T \bigl( f_t(x^*) + \varphi(x^*) \bigr) .
\end{align}

AdaGrad with composite mirror descent embodies one family of learning strategies for obtaining low regret. Although previously AdaGrad also has been presented with regularized dual averaging, in this work we restrict to composite mirror descent for simplicity. AdaGrad for composite mirror descent is described by the updates
\begin{align} \label{eqn:update}
x_{t+1} = \argmin_{x \in \X} 
              \left\{
                  \eta \langle g_t, x \rangle + \eta \varphi(x) + B_{\psi_t}(x, x_t) 
               \right\} ,
\end{align}
for some constant learning rate $\eta > 0$ and some adaptive choice of convex function $\psi_t$. 

We define $G_t := \sum_{s=1}^t g_s g_s^T$ and use the notation $\|x\|_A^2 = x^T A x$. 
In its full-matrix incarnation, AdaGrad uses the choice \citep{duchi2011adaptive}
\begin{align*}
\psi_t(x) = \frac{1}{2} \| x \|_{G_t^{1/2}}^2 .
\end{align*}
As shown by \cite{duchi2011adaptive} and reproduced in \cref{sec:regret} for convenience, full-matrix AdaGrad admits a strongly adaptive regret bound. Unfortunately, the update \eqref{eqn:update} for full-matrix AdaGrad is not tractable for large $n$ because it involves a matrix square root and solving an $n$-dimensional linear system, each of which costs time $O(n^3)$. In response to this issue, there is a diagonal version of AdaGrad that admits updates in time $O(n)$. For a square matrix $A$, let $\diag(A)$ be the diagonal matrix satisfying $\diag(A)_{ii} = A_{ii}$ for all $i$ and $\diag(A)_{ij} = 0$ for all $(i,j)$ such that $i \neq j$. The diagonal version of AdaGrad uses the choice \citep{duchi2011adaptive}
\begin{align*}
\psi_t(x) = \frac{1}{2} \| x \|_{\diag(G_t)^{1/2}}^2 .
\end{align*}
This method also admits theoretical guarantees (cf.~\citep{duchi2011adaptive} or \cref{sec:regret}). However, by design the diagonal method ignores the correlations between the components of the gradients.

\paragraph{CompAdaGrad.}
In this work, we introduce \emph{CompAdaGrad}, a method which combines the full-matrix approach in a subspace plus the diagonal approach in the complementary subspace. 

The idea of CompAdaGrad starts by restricting full-matrix AdaGrad's Bregman divergence to a low dimensional subspace by way of a mapping $\Pi: \real^n \rightarrow \real^k$ for some $k \leq n$ (and typically $k \ll n$). Since $\Pi G_t \Pi^T = \sum_{s=1}^T \Pi g_s (\Pi g_s)^T$, this leads to the modified Bregman divergence term
\begin{align} \label{eqn:radagrad}
\frac{1}{2} \| \Pi (x - x_t) \|_{(\Pi G_t \Pi^T)^{1/2}}^2 
= \frac{1}{2} \| x - x_t \|_{\Pi^T (\Pi G_t \Pi)^{1/2} \Pi}^2 ,
\end{align}
which was also used by \cite{krummenacher2014radagrad}.\footnote{Those authors select $\Pi$ to be an SRHT, described below, and we do this as well.} 
Our first remark is that if $k = n$ and $\Pi$ is in the orthogonal group, then it is easy to see that setting $\psi_t$ as in the RHS of \eqref{eqn:radagrad} recovers full-matrix AdaGrad. 

A deficiency of \eqref{eqn:radagrad} is that it ignores all of the action in the orthogonal complement of the image of $\Pi$; however, this action can be addressed naturally by directly considering the action in this complementary subspace. To this end, define $P$ to be the orthogonal projector corresponding to $\Pi$, defined as $P := \Pi^T (\Pi \Pi^T)^{-1} \Pi$. The corresponding complementary orthogonal projector is then $\Pperp := I - P$. Incorporating complementary regularization into the Bregman divergence with a diagonal approximation (since the dimension of the complementary subspace $n - k$ is presumably high) yields
\begin{align} \label{eqn:radagrad-complementary}
\frac{1}{2} \| \Pi (x - x_t) \|_{(\Pi G_t \Pi^T)^{1/2}}^2 
+ \frac{\tau}{2} \| \Pperp (x - x_t) \|_{\diag(G^\perp_t)^{1/2}}^2 ,
\end{align}
where $G^\perp_t := \sum_{s=1}^t (\Pperp g_s) (\Pperp g_s)^T = \Pperp G_T \Pperp$ and $\tau \geq 0$ is a parameter.

Note that when $k = n$ and $\Pi$ is in the orthogonal group, \eqref{eqn:radagrad-complementary} still recovers full-matrix AdaGrad (since the complementary subspace is empty). Additionally, when $k = 0$ and $\tau = 1$, \eqref{eqn:radagrad-complementary} recovers diagonal AdaGrad since $\Pperp = I$. Modulo a small modification to ensure strong convexity of $\psi_t$, we have just derived the method that we call CompAdaGrad, which is characterized by setting $\psi_t$ as
\begin{align} \label{eqn:psi-rp-oc}
\psi_t(x) = \frac{1}{2} \| x \|_{A^{(r)}_t + \tau A^{(c)}_t}^2 
\end{align}
\begin{flalign*}
\text{for} && 
A^{(r)}_t := \Pi^T \left( \Pi G_t \Pi^T  + \delta_r I \right)^{1/2} \Pi &&
\text{and} &&
A^{(c)}_t := 
\Pperp 
 \left( \diag \left( G^\perp_t \right)^{1/2} + \delta_c I \right) 
\Pperp . &&
\end{flalign*}

\paragraph{SRHT specialization.}
Throughout the rest of this paper, we take $\Pi$ to be a Subsampled Randomized Hadamard Transform (SRHT), defined as $\Pi: = \sqrt{\frac{n}{k}} R H \Sigma$ for a row selector $R \in \real^{k \times n}$, an $n$-dimensional (orthogonal) Walsh-Hadamard matrix $H$, and a diagonal Rademacher matrix $\Sigma \in \real^{n \times n}$ (whose diagonal entries are drawn i.i.d.~ as $+1$ and $-1$ with equal probability). The row selector $R$ is induced from the distribution of all cardinality-$k$ subsets of $n$ indices. 
The idea behind \eqref{eqn:psi-rp-oc} is that if there is interesting action in a low-dimensional subspace, the SRHT (i.e.~the $A^{(r)}$-part of the regularization) potentially can capture this action \citep{tropp2011improved}.
However, as the SRHT can only capture action in a low-dimensional subspace, it is critical to also include \emph{complementary} regularization by way of the $A^{(c)}$ part of the regularization. This is especially true in the adversarial regime, where an adversary can pick up on Learner's fixed SRHT and make all of the interesting action happen in the complement. In this sense, CompAdaGrad can have the benefit of full-matrix AdaGrad under an oblivious, easy adversary who restricts most of the action in a low-dimensional subspace, while also maintaining guarantees against a harder adversary who uses their knowledge of $\image(P)$.

The choices of $k$ and $\tau$ offer useful degrees of freedom for massive datasets. As $k$ increases toward $n$, the method more closely resembles full-matrix AdaGrad, and the regret bound generally becomes stronger as $k$ increases, while the computational complexity for various methods also becomes larger with larger $k$. The parameter $\tau$ on the other hand allows one to modulate how much emphasis to place on the complementary subspace, with this action increasingly ignored as $\tau$ decreases to 0. The next two sections help provide an understanding of the trade-off between regret and computation respectively.

\section{Regret Bounds}
\label{sec:regret}

Diagonal AdaGrad inherently is unable to adapt to correlations in a data sequence, as can be seen by Duchi et al.'s regret bound for this method (cf.~Theorem 5 of \cite{duchi2011adaptive}):
\begin{theorem} \label{thm:diagonal-regret}
\begin{align} \label{eqn:diagonal-regret}
\regret(f_1, \ldots, f_T) 
\leq \frac{1}{2 \eta} \max_{t \in [T]} \|x^* - x_t\|_\infty^2 
                     \sum_{j=1}^n \left( \sum_{t=1}^T g_{t,j}^2 \right)^{1/2} 
                     + \eta \sum_{j=1}^n \left( \sum_{t=1}^T g_{t,j}^2 \right)^{1/2} 
\end{align}
\end{theorem}
Although diagonal AdaGrad can perform well on high-dimensional sparse data, the above summations over the $n$ dimensions are symptomatic of its inability to adapt to highly correlated dimensions.

In contrast, full-matrix AdaGrad admits the regret bound (cf.~Theorem 7 of \cite{duchi2011adaptive})
\begin{theorem} \label{thm:full-regret}
\begin{align} \label{eqn:full-regret}
\regret(f_1, \ldots, f_T) 
\leq \frac{\delta}{\eta} \|x^*\|_2^2
                     + \frac{1}{2 \eta} \max_{t \in [T]} \|x^* - x_t\|_2^2 \tr(G_T^{1/2})
                     + \eta \tr(G_T^{1/2}) .
\end{align}
\end{theorem}
The above result, depending primarily on the trace of the covariance of the gradients, can exploit high correlations among the dimensions of points in a data sequence. We will show that, in certain situations of interest, CompAdaGrad can come close to the above regret guarantee at a small fraction of the computational complexity.

By design, it is quite straightforward to work out a regret bound for CompAdaGrad by leveraging the existing analysis of both full-matrix and diagonal AdaGrad from \cite{duchi2011adaptive}. Although in the context of concentration inequalities it is important to use the SRHT scaled as $\Pi = \sqrt{\frac{n}{k}} R H \Sigma$, in our analysis we instead analyze a variant of CompAdaGrad that is defined by replacing $\Pi$ with the unscaled $\tPi := R H \Sigma$ (which notably satisfies $\tPi \tPi^T = I$). The following regret bound is for this variant of CompAdaGrad. Also, we set $\delta_r = \delta_c = \delta$ because it simplifies the presentation. 
\begin{theorem} \label{thm:compressed-regret}
CompAdaGrad with learning rate $\eta > 0$ and $\delta > 0$ satisfies
\begin{align} \label{eqn:compressed-regret}
\regret(f_1, \ldots, f_T) 
\leq 
&\frac{\delta}{2 \eta} \|x^* - x_1\|_2^2 \nonumber \\
&\frac{1}{2 \eta} 
        \left( 
            \max_{t \in [T]} \| P (x^* - x_t) \|_2^2 
            \tr \left( (\tPi G_T \tPi^T)^{1/2} \right) 
            + \max_{t \in [T]} \| \Pperp (x^* - x_t) \|_\infty^2 \| Z^\perp \|_{2,1} 
        \right) \nonumber \\
       &+ \eta \left( 
               \tr \left( (\tPi G_T \tPi^T)^{1/2} \right) 
               + \| Z^\perp \|_{2,1} 
           \right) ,
\end{align}
where $Z^\perp = (\Pperp g_1 \, \ldots \, \Pperp g_T)^T$ and $A \mapsto \| A \|_{2,1}$ is the sum of the $\ell_2$ norms of the columns of $A$.
\end{theorem}
The proof can be found in \cref*{appendix:regret}.

Note that in the case where the data occupies a low-dimensional subspace and the SRHT preserves the action within this subspace, the above bound is similar to the regret bound for full-matrix AdaGrad \eqref{eqn:full-regret}. However, even when the data sequence is not so easy, \cref{thm:compressed-regret} still offers a fall-back guarantee based on the action in the orthogonal complement. In the event that the dimensions of the gradients are uncorrelated and the gradients do not occupy a low-dimensional subspace, the guarantee for diagonal AdaGrad provided by \cref{thm:diagonal-regret} could be better than the guarantee for CompAdaGrad provided by \cref{thm:compressed-regret}.

\section{Computations}
\label{sec:computations}

In this section, we show two important composite regularizers for which it is possible to compute the updates steps for CompAdaGrad: the squared $\ell_2$ regularizer and the $\ell_1$ regularizer. 
Some of the results below rely upon the following conjecture which we hope to affirm in the long version.
\begin{conjecture} \label{conjecture:k-sparse}
The $n$-dimensional Walsh-Hadamard Transform of an $r$-sparse vector can be computed in time $O(n \log r)$.
\end{conjecture}
For any results that rely on the conjecture for some variable (e.g.~$r$) indicating the sparsity level, we present the results instead with a \emph{primed} version of the variable (e.g.~$r'$) with the understanding that $r' = r$ if the conjecture is true and $r' = n$ otherwise.

Regardless of the veracity of \cref{conjecture:k-sparse}, we do however prove the following weaker result.
\begin{theorem} \label{thm:1-sparse}
The $n$-dimensional Walsh-Hadamard Transform of a $1$-sparse vector can be computed in time $O(n)$.
\end{theorem}
The proof is constructive, and the algorithm and its analysis can be found in \cref*{appendix:1-sparse-proof}.

\subsection{CompAdaGrad with the squared $\ell_2$ composite regularizer}

An update for squared $\ell_2$ composite mirror descent with the compressed Bregman divergence can be written in the form
\begin{align*}
\min_x \eta \langle g_t, x \rangle 
            + \frac{1}{2} \left\| \Pi (x- x_t) \right\|_{K_t}^2 
            + \frac{\tau}{2} \left\| \Pperp (x - x_t) \right\|_{D_t}^2 
            + \frac{\eta \lambda}{2} \|x\|_2^2 ,
\end{align*}
where $K_t = (\Pi G_t \Pi^T + \delta_r I)^{1/2}$ and $D_t = \diag(G^\perp_t)^{1/2} + \delta_c I$.

To simplify, we define $g := \eta g_t$, $K := K_t$, and $D = \tau D_t$, and we replace $\eta \lambda$; the above is then
\begin{align} \label{eqn:squared-ell2-simplified}
\min_x \langle g, x \rangle 
            + \frac{1}{2} \left\| \Pi (x- x_t) \right\|_K^2 
            + \frac{1}{2} \left\| \Pperp (x - x_t) \right\|_D^2 
            + \frac{\lambda}{2} \|x\|_2^2 .
\end{align}
Since $P$ is the orthogonal projector corresponding to the low-dimensional mapping $\Pi$, the above can be written equivalently as
\begin{align*}
\min_x \left\{
  \textcolor{dark-blue}{
    \langle P g, x \rangle 
    + \frac{1}{2} \left\| P (x- x_t) \right\|_{\Pi^T K \Pi}^2 
    + \frac{\lambda}{2} \|P x\|_2^2
  } 
  + \textcolor{dark-red}{
    \langle \Pperp g, x \rangle 
    + \frac{1}{2} \left\| \Pperp (x - x_t) \right\|_D^2
    + \frac{\lambda}{2} \|\Pperp x\|_2^2
  }
\right\} .
\end{align*}

From the above final rewriting, it is clear that the optimization decouples over the two subspaces, with the minimizer of the above problem $x^*$ being equal to the sum of
\begin{align} \label{eqn:problem-parallel}
&&& x^*_\parallel := 
  \argmin_{x \in \image(P)} \left\{ 
    \textcolor{dark-blue}{
       \langle P g, x \rangle 
      + \frac{1}{2} \left\| x - P x_t \right\|_{\Pi^T K \Pi}^2 
      + \frac{\lambda}{2} \|x\|_2^2
    }
\right\} && \\
\text{and} &&& x^*_\perp :=
\argmin_{x \in \image(\Pperp)} \left\{ 
  \textcolor{dark-red}{
    \langle \Pperp g, x \rangle 
    + \frac{1}{2} \left\| x - \Pperp x_t \right\|_D^2
    + \frac{\lambda}{2} \|x\|_2^2 
   }
    \right\} . && \label{eqn:problem-perp}
\end{align}

As shown in \cref*{appendix:computational}, the solution to \eqref{eqn:problem-parallel} is
\begin{align}
x^*_\parallel = (\Pi^T K \Pi + \lambda I)^{-1} (\Pi^T K \Pi x_t - P g) . \label{eqn:unconstrained-constrained}
\end{align}
This can be computed efficiently as follows; the proof can be found in \cref*{appendix:computational}.
\begin{lemma} \label{lemma:computation-squared-ell2-parallel}
$x^*_\parallel $ can be computed as
\begin{align} \label{eqn:computation-squared-ell2-parallel}
x^*_\parallel 
&= \Pi^T \left( \frac{n}{k} K + \lambda I \right)^{-1} \left( K \Pi x_t - \frac{k}{n} \Pi g \right)
\end{align}
in time $O(n \log k' + k^3)$.
\end{lemma}

To solve \eqref{eqn:problem-perp}, we take the dual. First, observe that \eqref{eqn:problem-perp} can be rewritten as
\begin{align*}
\begin{aligned}
&\min_x 
& & \langle \Pperp g, x \rangle 
       + \frac{1}{2} \left\| x - \Pperp x_t \right\|_D^2
       + \frac{\lambda}{2} \|x\|_2^2 \\
&\,\,\text{s.t.} 
& & \Pi x = \mathbf{0} .
\end{aligned}
\end{align*}

The Lagrangian is
\begin{align*}
\mathcal{L}(x, \nu) = 
  \langle \Pperp g, x \rangle 
  + \frac{1}{2} \left\| x - \Pperp x_t \right\|_D^2
  + \frac{\lambda}{2} \|x\|_2^2 
  + \langle \nu, \Pi x \rangle .
\end{align*}

\begin{lemma} \label{lemma:computation-squared-ell2-perp}
Problem \eqref{eqn:problem-perp} can be solved as
\begin{align} \label{eqn:x-in-terms-of-nu}
x^*_\perp = (D + \lambda I)^{-1} (-\Pperp g + D \Pperp x_t - \Pi^T \nu^*) 
\end{align}
with
\begin{align} \label{eqn:computation-for-nu}
\nu^* = 
\left( \Pi B \Pi^T \right)^{-1} 
\Pi B (-\Pperp g + D \Pperp x_t - D B y - \lambda B y + y) ,
\end{align}
for $B = (D + \lambda I)^{-1}$ and $y = -\Pperp g + D \Pperp x_t$. 
These computations can be completed in time $O(n k \log k)$.
\end{lemma}
The proof can be found in \cref*{appendix:computational}.

\subsection{CompAdaGrad with the $\ell_1$ composite regularizer}

We first sketch the LARS algorithm \citep{efron2004least}. We then show how the computations can be done efficiently for CompAdaGrad.

\paragraph{Re-expression as a LASSO problem.}

Using the same simplifications as in \eqref{eqn:squared-ell2-simplified}, the update step is
\begin{align*}
\argmin_x \langle g, x \rangle 
            + \frac{1}{2} \left\| x- x_t \right\|_A^2 
            + \lambda \|x\|_1 .
\end{align*}
for the symmetric matrix $A = \Pi^T K \Pi + \Pperp D \Pperp$. The above can be re-expressed as
\begin{align} \label{eqn:l1-update}
  x_{t+1} = \argmin_x \left\{ 
                                      \langle u, x \rangle 
                                      + \frac{1}{2} \langle x, A x \rangle 
                                      + \lambda \|x\|_1 
                                  \right\} ,
\end{align}
for $u = g - A x_t$. Note that $A$ is positive definite whenever $\delta_r, \delta_c > 0$.

The problem \eqref{eqn:l1-update} can be expressed as a LASSO problem, as the optimal objective value is equal to
\begin{align*}
\min_x \frac{1}{2} \|x\|_A^2 - \langle -A^{-1} u, x \rangle_A + \frac{1}{2} \left\| -A^{-1} u \right\|_A^2 + \lambda \|x\|_1 
&\equiv 
\min_x \frac{1}{2} \left\| x - (-A^{-1} u) \right\|_A^2 + \lambda \|x\|_1 \\
&\equiv 
\min_x \frac{1}{2} \left\| A^{1/2} x - (-A^{-1/2} u) \right\|^2 + \lambda \|x\|_1 .
\end{align*}

\paragraph{Efficient computations}

The standard LARS algorithm takes as input a matrix of covariates $X$ and targets $y$. In our case, $X = A^{1/2}$ is too expensive to compute since $A \in \real^{n \times n}$, while $y = A^{-1/2} u$ is similarly too expensive to compute. Therefore, we cannot just run the usual LARS algorithm but instead have to find efficient ways to perform some of the algorithm's internal computations.

The two computations for LARS that need to be handled differently are the computation of 
(a) the correlation of the covariates with the current residual; 
and 
(b) the entries of the Gram matrix for doing Cholesky Insert operations.

For both of the above, a certain $k$-by-$k$ matrix $Q$ arises in the computations which can be pre-computed and re-used for the entire run of LARS. $Q$ is defined as 
$\frac{n}{k} K + R H \Sigma D \Sigma H R^T$, or equivalently as $\frac{n}{k} K + R H D H R^T$. The second term can computed in time $O(n k \log k)$, in parallel over columns, precisely like $\Pi B \Pi^T$ from \eqref{eqn:computation-for-nu}, as explained in the proof of \cref{lemma:computation-squared-ell2-perp}.

Given a current hypothesis $\beta$, the correlation of the covariates with the residual can be computed as $X^T(y - X \beta) = A^{1/2} A^{-1/2} u - A^{1/2} A^{1/2} \beta = u - A \beta$. Thus, a matrix vector multiplication is the main work. This operation can be completed in time $O(n \log k' + k^2)$, as shown in \cref*{appendix:matrix-vector}.

When performing a Cholesky Insert (increasing the active set $\Lambda$ by one), we need to compute part of a column of the Gram matrix. If the new dimension is $i$, we need to compute $G_{ii}$ and $G_{ij}$ for each $j \in \Lambda$. Since the Gram matrix $G = A$, we just need to compute $e_i^T A e_j$ for each $j \in \Lambda$. 

To compute $G_{ii}$, observe that
\begin{align} \label{eqn:Gram-ii}
G_{ii} = A_{ii} = d_i \langle R H e_i, Q R H e_i \rangle - 2 d_i \frac{k}{n} ,
\end{align}
To compute \eqref{eqn:Gram-ii}, we first compute $R H e_i$ in $O(k \log n)$ and store this for use in future rounds. The additional cost for computing $Q R H e_i$ is then $O(k^2)$, and we re-use this result in \eqref{eqn:Gram-column}. So the cost is $O(k \log n + k^2)$. 
To compute $G_{i,\Lambda}$, we compute for each $j \in \Lambda$
\begin{align} \label{eqn:Gram-column}
G_{ij} = \sigma_i \sigma_j \left( \langle R H e_j, Q R H e_i \rangle - (d_i + d_j) \langle R H e_j, R H e_i \rangle \right) .
\end{align}
Computing \eqref{eqn:Gram-column} can be done in $O(|\Lambda| k)$ since we have already stored $R H e_j$ for all $j \in \Lambda$ and already computed $Q R H e_i$.

\section{Experiments}
\label{sec:experiments}

For all the experiments we used the squared $\ell_2$ regularizer. For CompAdaGrad, we always set $\delta_r$ and $\delta_c$ to be equal and hereafter refer to them both as $\delta$. 
When reporting test error for MNIST and Reuters RCV1, we select the values of $\delta$, $\eta$, $\lambda$, and $\tau$ that attained the lowest online zero-one loss.

\begin{figure}
\begin{floatrow}
\ffigbox{
\includegraphics[width=75mm]{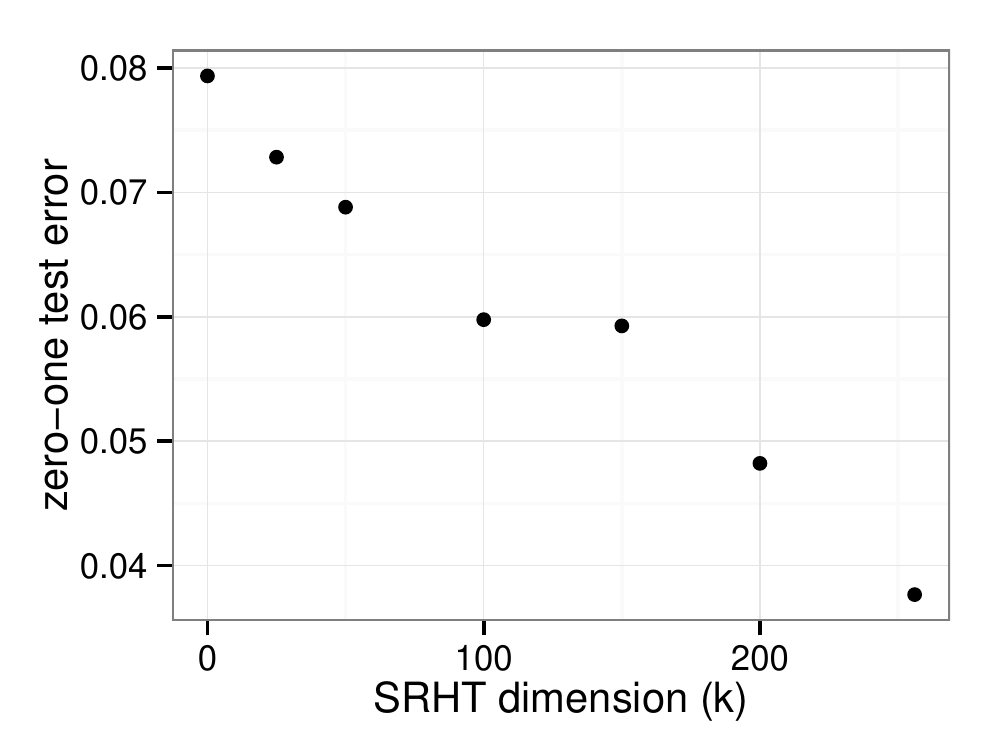}
}{
  \caption{\label{fig:mnist} MNIST 4 vs 9 test error.}
}
\ffigbox{
\includegraphics[width=40mm]{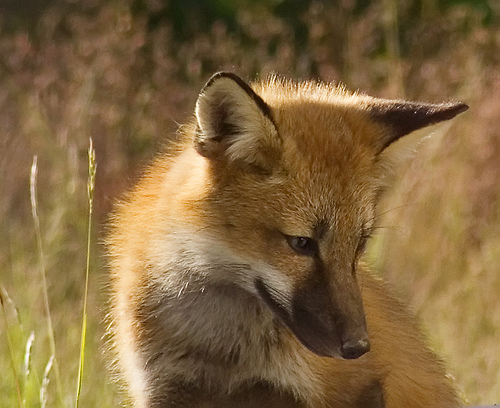}
\includegraphics[width=40mm]{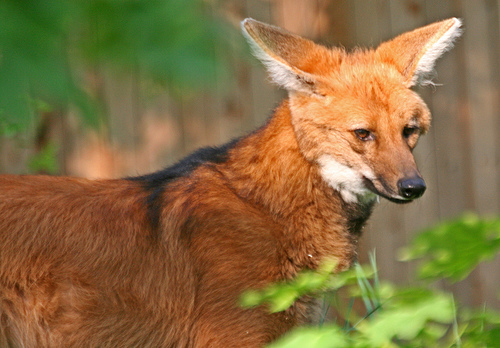}
}{
  \caption{A kit fox and a red wolf}
}
\end{floatrow}
\end{figure}

\paragraph{MNIST 4 vs 9.}
This dataset was constructed by selecting 400 random prototypes (200 from each class) from the training set and using as features a Gaussian kernel computation of each data point with each of the 400 prototypes. We train on the entire training set for MNIST 4 and 9 and test on the test set for MNIST 4 and 9. 
The results in \cref{fig:mnist} indicate that CompAdaGrad well-outperforms the diagonal method, and as the SRHT dimension increases from 25 to 256, the gains over the diagonal method are all the larger. It is worth mentioning that the CompAdaGrad improves over diagonal AdaGrad even for $k = 25$. We suspect that that CompAdaGrad surpasses diagonal AdaGrad on this task because the features are dense and highly correlated.

\paragraph{Reuters RCV1.}
We used Reuters RCV1-v2 \citep{lewis2004rcv1}. After stopping, stemming, and retaining only those unigrams and bigrams that occurred at least twice in the corpus, the data consisted of 800,000 documents in 1,889,478 dimensions. We computed gradients in mini-batches of 160 points and performed a hypothesis update from each resulting averaged gradient. We generated 4 random permutations of the data. For each permutation, we train on the first 75\% and test on the last 25\%. Each of the 4 runs is an experiment. 
We report the average test error over the 4 experiments. This test error need not correspond to a single $(\eta, \lambda,\delta,(\tau))$ configuration. 
As shown in \cref{table:reuters}, CompAdaGrad with $k=512$ does not improve upon the diagonal version. We suspect the lack of improvement is because the data is very sparse and does not admit a low-dimensional linear subspace capturing a large amount of the action in the data. Therefore, it does not help much to pick up on correlations.

\paragraph{Foxes and Wolves.}
We grabbed a visual-bag-of-words representation for 829 images of kit foxes (\emph{Vulpes macrotis}) and 1,156 images of red wolves (\emph{Canis rufus}) from ImageNet \citep{russakovsky2015imagenet} (originally from Flickr). The original features are 1000 visual words which represent 1000 clusters of a random subset of 10 million SIFT features. 
We computed visual word bigrams by selecting\footnote{We selected 0.002 based on the quantiles of the distance distribution of all pairs of visual words in images.} a radius of 0.002 for visual words to be considered collocated (based on $(x,y)$ coordinates for the visual words); this led to 238,822 bigrams which occurred at least once in the 1,985 images. The final dataset for learning consisted of the unigram counts and bigram counts, constituting 239,822 features.

We trained on the first 75\% of a random permutation of the data and tested on the remainder. 
Because the number of instances in this dataset is small relative to the dimension, the online zero-one loss is not a sensible rule for parameter tuning as a relatively large number hypotheses contributing to the online loss are from early rounds where not much learning has taken place. We therefore tuned the parameters based on the online training zero-one loss of the final hypothesis of the online learning algorithm. The results presented in \cref{table:foxes-wolves} are inconclusive. Both diagonal AdaGrad and CompAdaGrad with $k=256$ obtain the same zero-one test risk. This either could be due to the hardness of the task, the small sample size, or the fact that the features are still quite sparse due to the visual bag-of-words bigram representation.

\begin{figure}
\begin{floatrow}
\capbtabbox{
\begin{tabular}{c c}
\toprule
Method                   & Zero-one test risk \\
\midrule
Diagonal            &  0.056084 \\
CompAdaGrad-512  & 0.056155 \\ 
\bottomrule
\end{tabular}
}{
  \caption{\label{table:reuters} Reuters RCV1 test error.}
}
\capbtabbox{
\begin{tabular}{c c}
\toprule
Method      & Zero-one test risk \\
\midrule
Diagonal   &  0.2696 \\
CompAdaGrad-256 & 0.2696 \\ 
\bottomrule
\end{tabular}
}{
  \caption{\label{table:foxes-wolves}  Foxes and Wolves test error}
}
\end{floatrow}
\end{figure}

\section{Conclusion}
\label{sec:discussion}

We have introduced a new adaptive gradient method, CompAdaGrad, which can obtain regret bounds competitive with full-matrix AdaGrad under easier adversaries that put most of the interesting action in a low-dimensional subspace, while maintaining a fallback regret bound in the case that the adversary is not easy. CompAdaGrad also admits efficient updates for certain choices of the composite regularizer. In preliminary experiments, we demonstrate that CompAdaGrad can sometimes obtain lower risk as compared to diagonal AdaGrad, but this outcome appears to be tied to the density and more importantly the correlation of the features. In the future, we intend to apply CompAdaGrad to massive datasets with dense, highly correlated features. Along these lines, one promising application for future work is to incorporate CompAdaGrad into the training of deep learning methods.

\subsubsection*{Acknowledgments}

This work is supported in part by the Australian Research Council Discovery Project DP0987773 and NICTA, which is funded by the Australian Government through the Department of Communications and the Australian Research Council through the ICT Centre of Excellence Program.

\newpage

\renewcommand{\refname}{{\normalsize References}}

\bibliography{compadagrad}

\begin{thebibliography}{10}
\providecommand{\natexlab}[1]{#1}
\providecommand{\url}[1]{\texttt{#1}}
\expandafter\ifx\csname urlstyle\endcsname\relax
  \providecommand{\doi}[1]{doi: #1}\else
  \providecommand{\doi}{doi: \begingroup \urlstyle{rm}\Url}\fi

\bibitem[Abernethy et~al.(2008)Abernethy, Bartlett, Rakhlin, and
  Tewari]{abernethy2008optimal}
Jacob Abernethy, Peter~L Bartlett, Alexander Rakhlin, and Ambuj Tewari.
\newblock Optimal strategies and minimax lower bounds for online convex games.
\newblock In \emph{Proceedings of the nineteenth annual conference on
  computational learning theory}, 2008.

\bibitem[Dean et~al.(2012)Dean, Corrado, Monga, Chen, Devin, Mao, Senior,
  Tucker, Yang, Le, et~al.]{dean2012large}
Jeffrey Dean, Greg Corrado, Rajat Monga, Kai Chen, Matthieu Devin, Mark Mao,
  Andrew Senior, Paul Tucker, Ke~Yang, Quoc~V Le, et~al.
\newblock Large scale distributed deep networks.
\newblock In \emph{Advances in Neural Information Processing Systems}, pages
  1223--1231, 2012.

\bibitem[Duchi et~al.(2011)Duchi, Hazan, and Singer]{duchi2011adaptive}
John Duchi, Elad Hazan, and Yoram Singer.
\newblock Adaptive subgradient methods for online learning and stochastic
  optimization.
\newblock \emph{Journal of Machine Learning Research}, 12:\penalty0 2121--2159,
  2011.

\bibitem[Efron et~al.(2004)Efron, Hastie, Johnstone, Tibshirani,
  et~al.]{efron2004least}
Bradley Efron, Trevor Hastie, Iain Johnstone, Robert Tibshirani, et~al.
\newblock Least angle regression.
\newblock \emph{The Annals of statistics}, 32\penalty0 (2):\penalty0 407--499,
  2004.

\bibitem[Gr\"unwald et~al.(2013)Gr\"unwald, Koolen, and
  Rakhlin]{grunwald2013learning}
Peter Gr\"unwald, Wouter~M. Koolen, and Alexander Rakhlin, editors.
\newblock \emph{NIPS Workshop on "Learning faster from easy data"}, 2013.

\bibitem[Krummenacher and McWilliams(2014)]{krummenacher2014radagrad}
Gabriel Krummenacher and Brian McWilliams.
\newblock {RadaGrad}: Random projections for adaptive stochastic optimization.
\newblock In \emph{OPT 2014: 7th NIPS Workshop on Optimization for Machine
  Learning}, 2014.

\bibitem[Lewis et~al.(2004)Lewis, Yang, Rose, and Li]{lewis2004rcv1}
David~D Lewis, Yiming Yang, Tony~G Rose, and Fan Li.
\newblock Rcv1: A new benchmark collection for text categorization research.
\newblock \emph{Journal of Machine Learning Research}, 5:\penalty0 361--397,
  2004.

\bibitem[Russakovsky et~al.(2015)Russakovsky, Deng, Su, Krause, Satheesh, Ma,
  Huang, Karpathy, Khosla, Bernstein, Berg, and
  Fei-Fei]{russakovsky2015imagenet}
Olga Russakovsky, Jia Deng, Hao Su, Jonathan Krause, Sanjeev Satheesh, Sean Ma,
  Zhiheng Huang, Andrej Karpathy, Aditya Khosla, Michael Bernstein,
  Alexander~C. Berg, and Li~Fei-Fei.
\newblock {ImageNet Large Scale Visual Recognition Challenge}.
\newblock \emph{International Journal of Computer Vision (IJCV)}, 2015.
\newblock \doi{10.1007/s11263-015-0816-y}.

\bibitem[Tropp(2011)]{tropp2011improved}
Joel~A Tropp.
\newblock Improved analysis of the subsampled randomized hadamard transform.
\newblock \emph{Advances in Adaptive Data Analysis}, 3\penalty0
  (01n02):\penalty0 115--126, 2011.

\bibitem[Zinkevich(2003)]{zinkevich2003online}
Martin Zinkevich.
\newblock Online convex programming and generalized infinitesimal gradient
  ascent.
\newblock In \emph{AAAI}, 2003.

\end{thebibliography}

\newpage

\appendix

\section{Proofs}

\subsection{Proof of regret bound}
\label{appendix:regret}

\begin{proof}[Proof of \cref{thm:compressed-regret}]
We begin similar to Proposition 3 of \cite{duchi2011adaptive}:
\begin{align*}
&\eta \left( f_t(x_t) + \varphi(x_{t+1}) - f_t(x^*) - \varphi(x^*) \right) \\
&\leq \textcolor{dark-green}{B_{\psi_t}(x^*, x_t) - B_{\psi_t}(x^*, x_{t+1})} 
          \textcolor{red}{- B_{\psi_t}(x_{t+1}, x_t) + \eta \langle \eta^{-1/2} (x_t - x_{t+1}), \eta^{1/2} g_t \rangle} .
\end{align*}

In our case, $\psi_t$ is defined as
\begin{align}
\psi_t(x) 
:= \frac{1}{2} \| x \|^2_{\tPi^T ((\tPi G_t \tPi^T)^{1/2} + \delta I) \tPi}
   + \frac{1}{2} \| x \|^2_{\Pperp (D_t^{1/2} + \delta I) \Pperp} ,  \label{eqn:recall-psi}
\end{align}
with $D_t$ is the diagonal matrix defined by the diagonal entries of $\Pperp G_t \Pperp$. 

We bound the red and green parts in turn. For readability, we define $\tilde{G_t} := \tPi G_t \tPi^T$.

\paragraph{\textcolor{red}{Red part}}
The red part can be rewritten as
\begin{align*}
&-\frac{1}{2} \| \tPi P (x_t - x_{t+1}) \|_{\tilde{G}_t^{1/2} + \delta I}^2 
+ \eta \langle \eta^{-1/2} P (x_t - x_{t+1}), \eta^{1/2} P g_t \rangle \\
&-\frac{1}{2} \| \Pperp (x_t - x_{t+1}) \|_{D_t^{1/2} + \delta I}^2 
+ \eta \langle \eta^{-1/2} \Pperp (x_t - x_{t+1}), \eta^{1/2} \Pperp g_t \rangle ,
\end{align*}
which it is easy to verify is equal to
\begin{align*}
&-\frac{1}{2} \| \tPi P (x_t - x_{t+1}) \|_{\tilde{G}_t^{1/2} + \delta I}^2 
+ \eta \langle \eta^{-1/2} \tPi P (x_t - x_{t+1}), \eta^{1/2} \tPi P g_t \rangle \\
&-\frac{1}{2} \| \Pperp (x_t - x_{t+1}) \|_{D_t^{1/2} + \delta I}^2 
+ \eta \langle \eta^{-1/2} \Pperp (x_t - x_{t+1}), \eta^{1/2} \Pperp g_t \rangle 
\end{align*}
The above can bounded from above by the Fenchel-Young inequality as
\begin{align*}
&-\frac{1}{2} \| \tPi P (x_t - x_{t+1}) \|_{\tilde{G}_t^{1/2} + \delta I}^2 
+ \frac{1}{2} \| \tPi P (x_t - x_{t+1}) \|_{\tilde{G}_t^{1/2} + \delta I}^2 
+ \frac{\eta^2}{2} \| \tPi P g_t \|_{(\tilde{G}_t^{1/2} + \delta I)^{-1}}^2 \\
&-\frac{1}{2} \| \Pperp (x_t - x_{t+1}) \|_{D_t^{1/2} + \delta I}^2 
+ \frac{1}{2} \| \Pperp (x_t - x_{t+1}) \|_{D_t^{1/2} + \delta I}^2 
+ \frac{\eta^2}{2} \| \Pperp g_t \|_{(D_t^{1/2} + \delta I)^{-1}}^2 ,
\end{align*}
which is just
\begin{align*}
\frac{\eta^2}{2} \| \tPi g_t \|_{(\tilde{G}_t^{1/2} + \delta I)^{-1}}^2 
+ \frac{\eta^2}{2} \| \Pperp g_t \|_{(D_t^{1/2} + \delta I)^{-1}}^2 .
\end{align*}

Summing the bound due to the red part for $t = 1$ to $T$ yields the following two bounds: \\
First, from Lemma 10 of \cite{duchi2011adaptive} we have
\begin{align*}
\sum_{t=1}^T \| \tPi g_t \|_{(\tilde{G}_t^{1/2} + \delta I)^{-1}}^2 
\leq \sum_{t=1}^T \| \tPi g_t \|_{\tilde{G}_t^{-1/2}}^2 
\leq 2 \tr \left( (\tPi G_T \tPi^T )^{1/2} \right) .
\end{align*}
Next, from Lemma 4 of \cite{duchi2011adaptive} we have
\begin{align*}
\sum_{t=1}^T \| \Pperp g_t \|_{(D_t^{1/2} + \delta I)^{-1}}^2 
\leq \sum_{t=1}^T \| \Pperp g_t \|_{D_t^{-1/2}}^2 
\leq 2 \sum_{j=1}^n \left( \sum_{t=1}^T [\Pperp g_t]_j^2 \right)^{1/2} .
\end{align*}

Having sufficiently bounded the red part, we now turn to the green part.

\paragraph{\textcolor{dark-green}{Green part}}
We again sum from $t = 1$ to $T$, yielding
\begin{align}
&\sum_{t=1}^T \left( B_{\psi_t}(x^*, x_t) - B_{\psi_t}(x^*, x_{t+1}) \right) \nonumber \\
&= B_{\psi_1}(x^*, x_1) 
    + \sum_{t=1}^{T-1} B_{\psi_{t+1}}(x^*, x_{t+1})
    - \sum_{t=1}^T B_{\psi_t}(x^*, x_{t+1}) \nonumber \\
&\leq B_{\psi_1}(x^*, x_1) 
          + \sum_{t=1}^{T-1} \left( B_{\psi_{t+1}}(x^*, x_{t+1}) - B_{\psi_t}(x^*, x_{t+1}) \right) . \label{eqn:general-green}
\end{align}

Recalling the form of $\psi_t$ in \eqref{eqn:recall-psi} which has a ``$P$'' part and ``$\Pperp$'' part, we can decompose our analysis of \eqref{eqn:general-green} similarly. 

We first analyze the ``$P$'' part of \eqref{eqn:general-green}:
\begin{align*}
&\frac{1}{2} \| P (x^* - x_1) \|_{\tPi^T (\tilde{G}_1^{1/2} + \delta I) \tPi}^2 \\
&\quad+ \frac{1}{2} \sum_{t=1}^{T-1} 
                \left( 
                    \| P (x^* - x_{t+1}) \|_{\tPi^T (\tilde{G}_{t+1}^{1/2} + \delta I) \tPi}^2
                    - \| P (x^* - x_{t+1}) \|_{\tPi^T (\tilde{G}_t^{1/2} + \delta I) \tPi}^2
                 \right) \\
&= \frac{1}{2} \| P (x^* - x_1) \|_{\tPi^T (\tilde{G}_1^{1/2} + \delta I) \tPi}^2 \\
&\quad+ \frac{1}{2} \sum_{t=1}^{T-1} 
                \left( 
                    \| P (x^* - x_{t+1}) \|_{\tPi^T \tilde{G}_{t+1}^{1/2} \tPi}^2
                    - \| P (x^* - x_{t+1}) \|_{\tPi^T \tilde{G}_t^{1/2} \tPi}^2
                 \right) \\
&\leq \frac{1}{2} \| P (x^* - x_1) \|_{\tPi^T (\tilde{G}_1^{1/2} + \delta I) \tPi}^2 \\
&\quad+ \frac{1}{2} \sum_{t=1}^{T-1} 
           \| P (x^* - x_{t+1}) \|_2^2 \lambda_{\max} \left( \tPi^T \bigl( \tilde{G}_{t+1}^{1/2} - \tilde{G}_t^{1/2} \bigr) \tPi \right) \\
&\leq \frac{1}{2} \| P (x^* - x_1) \|_{\tPi^T (\tilde{G}_1^{1/2} + \delta I) \tPi}^2 \\
&\quad+ \frac{1}{2} \sum_{t=1}^{T-1} 
           \| P (x^* - x_{t+1}) \|_2^2 \tr \left( \tPi^T \bigl( \tilde{G}_{t+1}^{1/2} - \tilde{G}_t^{1/2} \bigr) \tPi \right) \\
&\leq \frac{1}{2} \| P (x^* - x_1) \|_{\tPi^T (\tilde{G}_1^{1/2} + \delta I) \tPi}^2 \\
&\quad+ \frac{1}{2} \max_{t \in [T]} \| P (x^* - x_t) \|_2^2
      \sum_{t=1}^{T-1} 
           \tr \left( \tPi^T \bigl( \tilde{G}_{t+1}^{1/2} - \tilde{G}_t^{1/2} \bigr) \tPi \right) \\
&\leq \frac{1}{2} \| P (x^* - x_1) \|_{\tPi^T (\tilde{G}_1^{1/2} + \delta I) \tPi}^2 \\
&\quad+ \frac{1}{2} \max_{t \in [T]} \| P (x^* - x_t) \|_2^2 
                \tr \left( \tPi^T \tilde{G}_T^{1/2} \tPi \right) 
                - \frac{1}{2} \| P (x^* - x_1) \|_2^2 
                \tr \left( \tPi^T \tilde{G}_1^{1/2} \tPi \right) \\
&\leq \frac{\delta}{2} \| P(x^* - x_1) \|_2^2
      + \frac{1}{2} \max_{t \in [T]} \| P (x^* - x_t) \|_2^2 
                \tr \left( \tPi^T \tilde{G}_T^{1/2} \tPi \right) \\
&= \frac{\delta}{2} \| P(x^* - x_1) \|_2^2 
      + \frac{1}{2} \max_{t \in [T]} \| P (x^* - x_t) \|_2^2 
                \tr \left( \tilde{G}_T^{1/2} \tPi \tPi^T \right) \\
&= \frac{\delta}{2} \| P(x^* - x_1) \|_2^2 
      + \frac{1}{2} \max_{t \in [T]} \| P (x^* - x_t) \|_2^2 
                \tr \left( \tilde{G}_T^{1/2} \right) ,
\end{align*}
where the last inequality follows because $\tPi^T \tPi = P$, 
and the last equality follows since $\tPi \tPi^T = I$.

We now turn to the ``$\Pperp$ part of \eqref{eqn:general-green}:
\begin{align*}
&\frac{1}{2} \| \Pperp (x^* - x_1) \|_{D_1^{1/2} + \delta I}^2 \\
&\quad+ \frac{1}{2} \sum_{t=1}^{T-1} 
                  \left(
                      \| \Pperp (x^* - x_{t+1}) \|_{D_{t+1}^{1/2} + \delta I}^2 
                      - \| \Pperp (x^* - x_{t+1}) \|_{D_t^{1/2} + \delta I}^2
                   \right) \\
&= \frac{1}{2} \| \Pperp (x^* - x_1) \|_{D_1^{1/2} + \delta I}^2 \\
&\quad+ \frac{1}{2} \sum_{t=1}^{T-1} 
                  \left\langle \Pperp (x^* - x_{t+1}) , 
                                      (D_{t+1}^{1/2} - D_t^{1/2}) \Pperp (x^* - x_{t+1}) 
                  \right \rangle \\
&\leq \frac{1}{2} \| \Pperp (x^* - x_1) \|_{D_1^{1/2} + \delta I}^2 \\
&\quad+ \frac{1}{2} \sum_{t=1}^{T-1} 
                  \| \Pperp (x^* - x_t) \|_\infty^2 
                  \langle D_{t+1}^{1/2} - D_t^{1/2} , \mathbf{1} \rangle \\
&\leq \frac{1}{2} \| \Pperp (x^* - x_1) \|_{D_1^{1/2} + \delta I}^2 \\
&\quad+ \frac{1}{2} \max_{t \in [T]} \| \Pperp (x^* - x_t) \|_\infty^2 
                \sum_{t=1}^{T-1} 
                \langle D_{t+1}^{1/2} - D_t^{1/2} , \mathbf{1} \rangle \\
&\leq \frac{1}{2} \| \Pperp (x^* - x_1) \|_{D_1^{1/2} + \delta I}^2 \\
&\quad+ \frac{1}{2} \max_{t \in [T]} \| \Pperp (x^* - x_t) \|_\infty^2  
                \langle D_T^{1/2} , \mathbf{1} \rangle 
                - \frac{1}{2} \| \Pperp (x^* - x_1) \|_\infty^2
                   \langle D_1^{1/2} , \mathbf{1} \rangle \\
&\leq \frac{\delta}{2} \| \Pperp (x^* - x_1) \|_2^2
          + \frac{1}{2} \max_{t \in [T]} \| \Pperp (x^* - x_t) \|_\infty^2 
                \langle D_T^{1/2} , \mathbf{1} \rangle .
\end{align*}

\paragraph{The final regret bound}
Putting everything above together yields the final bound
\begin{align*}
&\sum_{t=1}^T \bigl( f_t(x_t) + \varphi(x_{t+1}) - f_t(x^*) - \varphi(x^*) \bigr) \\
&\leq \frac{\delta}{2 \eta} \|x^* - x_1\|_2^2 \\
&\quad + \frac{1}{2 \eta} 
             \left( 
                 \max_{t \in [T]} \| P (x^* - x_t) \|_2^2 
                 \tr \left( (\tPi G_T \tPi)^{1/2} \right) 
                 + \max_{t \in [T]} \| \Pperp (x^* - x_t) \|_\infty^2 
                     \langle D_T^{1/2}, \mathbf{1} \rangle 
             \right) \\
&\quad + \eta \left( 
               \tr \left( \tilde{G}_T^{1/2} \right) 
               + \langle D_T^{1/2}, \mathbf{1} \rangle 
           \right) 
\end{align*}
with
\begin{align*}
\langle D_T^{1/2} , \mathbf{1} \rangle 
= \sum_{j=1}^n \left( \sum_{t=1}^T [\Pperp g_t]_j^2 \right)^{1/2} .
\end{align*}
\end{proof}

\subsection{Proof of computational results}
\label{appendix:computational}

First, we establish that \eqref{eqn:unconstrained-constrained} is in fact the solution to the constrained problem \eqref{eqn:problem-parallel}.

We rely on the following claim:
\begin{align}
(\Pi^T K \Pi + \lambda I)^{-1} \Pi^T = \Pi^T (K \Pi \Pi^T + \lambda I)^{-1} . \label{eqn:claim}
\end{align}
To see this, multiply both sides on the left by $(\Pi^T K \Pi + \lambda I)$, yielding
\begin{align*}
\Pi^T 
&= (\Pi^T K \Pi + \lambda I) \Pi^T (K \Pi \Pi^T + \lambda I)^{-1} \\
&= \Pi^T (K \Pi \Pi^T + \lambda I) (K \Pi \Pi^T + \lambda I)^{-1} 
= \Pi^T .
\end{align*}

Now, observe that \eqref{eqn:unconstrained-constrained} is the solution to the unconstrained version of \eqref{eqn:problem-parallel}. Thus, it is sufficient to show that $\Pperp (\Pi^T K \Pi + \lambda I)^{-1} (\Pi^T K \Pi x_t - P g)$ is equal to zero. This is indeed true since
\begin{align*}
&\Pperp (\Pi^T K \Pi + \lambda I)^{-1} (\Pi^T K \Pi x_t - P g) \\
&= \left( I - \frac{k}{n} \Pi^T \Pi \right) (\Pi^T K \Pi + \lambda I)^{-1} \left( \Pi^T K \Pi x_t - \frac{k}{n} \Pi^T \Pi g \right) \\
&= \left( I - \frac{k}{n} \Pi^T \Pi \right) (\Pi^T K \Pi + \lambda I)^{-1} \Pi^T \left( K \Pi x_t - \frac{k}{n} \Pi g \right) \\
&= \left( I - \frac{k}{n} \Pi^T \Pi \right) \Pi^T (K \Pi \Pi^T + \lambda I)^{-1} \left( K \Pi x_t - \frac{k}{n} \Pi g \right) \\
&= \left( \Pi^T - \Pi^T \right) (K \Pi \Pi^T + \lambda I)^{-1} \left( K \Pi x_t - \frac{k}{n} \Pi g \right) \\
&= 0 .
\end{align*}

\begin{proof}[Proof of \cref{lemma:computation-squared-ell2-parallel}]
First, 
\begin{align}
(\Pi^T K \Pi + \lambda I)^{-1} (\Pi^T K \Pi x_t - P g) 
= (\Pi^T K \Pi + \lambda I)^{-1} \Pi^T \left( K \Pi x_t - \frac{k}{n} \Pi g \right) \label{eqn:midway}
\end{align}
since $P = \Pi^T (\Pi \Pi^T)^{-1} \Pi = \frac{k}{n} \Pi^T \Pi$.

Using \eqref{eqn:claim}, we have that the RHS of \eqref{eqn:midway} is equal to
\begin{align*}
\Pi^T (K \Pi \Pi^T + \lambda I)^{-1} \left( K \Pi x_t - \frac{k}{n} \Pi g \right) 
= \Pi^T \left( \frac{n}{k} K + \lambda I \right)^{-1} \left( K \Pi x_t - \frac{k}{n} \Pi g \right) .
\end{align*}
\end{proof}

\begin{proof}[Proof of \cref{lemma:computation-squared-ell2-perp}]

We first derive the expressions for $x^*_\perp$ and $\nu^*$ and then establish the computational complexity result.

\paragraph{Proof of expressions for $x^*_\perp$ and $\nu^*$.}
At the maximum the partial gradient of the Lagrangian WRT $x$ must be zero, and so
\begin{align*}
\frac{\partial \mathcal{L}}{\partial x} = \mathbf{0} = \Pperp g + D (x - \Pperp x_t) + \lambda x + \Pi^T \nu .
\end{align*}
Thus, we have the relation
\begin{align} \label{eqn:x-in-terms-of-nu-II}
x = (D + \lambda I)^{-1} (-\Pperp g + D \Pperp x_t - \Pi^T \nu) 
\end{align}
which can be computed in time $O(n \log k)$ assuming $O(n \log k)$ computation of the Walsh-Hadamard transform of a $k$-sparse vector.

Define $y:= -\Pperp g + D \Pperp x_t$ and $B := (D + \lambda I)^{-1}$. Plugging in this expression for $x$ into the Lagrangian yields the dual problem
\begin{align*}
\max_\nu 
  \left\langle \Pperp g, B (y - \Pi^T \nu) \right\rangle 
  + \frac{1}{2} \left\| B (y - \Pi^T \nu) - \Pperp x_t \right\|_D^2 
  + \frac{\lambda}{2} \left\| B (y - \Pi^T \nu) \right\|_2^2 
  + \left\langle \nu, \Pi B (y - \Pi^T \nu) \right\rangle .
\end{align*}
which is equivalent to the problem
\begin{align} \label{eqn:max-problem}
\max_\nu 
  -\left\langle \Pperp g, B \Pi^T \nu \right\rangle 
  + \frac{1}{2} \left\| B (y - \Pi^T \nu) - \Pperp x_t \right\|_D^2 
  + \frac{\lambda}{2} \left\| B (y - \Pi^T \nu) \right\|_2^2 
  + \left\langle \nu, \Pi B y \right\rangle 
  - \| \nu \|_{\Pi B \Pi^T}^2 . 
\end{align}

Let us verify that the objective is concave. The Hessian is
\begin{align*}
\Pi B D B \Pi^T + \lambda \Pi B^2 \Pi^T - 2 \Pi B \Pi^T 
&= \Pi \left( B D B + \lambda B^2 - 2 B \right) \Pi^T .
\end{align*}
The term on the RHS sandwiched between $\Pi$ and $\Pi^T$ expands to
\begin{align*}
(D + \lambda I)^{-2} D + \lambda (D + \lambda I)^{-2} - 2 (D + \lambda I)^{-1} ,
\end{align*}
which (by way of the diagonal structure) easily works out to be $-(D + \lambda I)^{-1}$. Hence, the problem is strongly concave. 

Thus, we have reduced the problem to a low-dimensional concave unconstrained problem which can be solved analytically. Differentiating \emph{just the first-order (in $\nu$) terms} of the objective of \eqref{eqn:max-problem} WRT $\nu$ yields
\begin{align*}
&\Pi B \Pperp g 
+ \Pi B D B y
- \Pi B D \Pperp x_t 
+ \lambda \Pi B^2 y 
- \Pi B y \\
&= \Pi B (\Pperp g + D B y - D \Pperp x_t + \lambda B y - y) .
\end{align*}

Finally, setting the gradient to zero and solving for $\nu$ yields
\begin{align} \label{eqn:computation-for-nu-II}
\nu = 
\left( \Pi (D + \lambda I)^{-1} \Pi^T \right)^{-1} 
\Pi B (-\Pperp g + D \Pperp x_t - D B y - \lambda B y + y) .
\end{align}

We can then compute $x$ using \eqref{eqn:x-in-terms-of-nu-II}.

\paragraph{Proof of computational complexity.}
First, the computation for \eqref{eqn:computation-for-nu} can be done efficiently because the vector to which the inverse is applied can be computed in $O(n \log k')$. To see this, observe that for $x \in \real^n$, we have
$P x = \Sigma H R^T R H \Sigma x$, which involves (in sequence) scaling by $\Sigma$, a trimmed Walsh-Hadamard Transform (WHT) in $O(n \log k)$, applying $R^T$ to create a $k$-sparse $n$-dimensional vector in $O(n)$, application of a WHT to this $k$-sparse vector in $O(n \log k')$, and a final scaling by $\Sigma$.

Computing the linear system matrix (involved in the inverse) is the most expensive step: the columns of this $k$-by-$k$ matrix can be computed in parallel, and each column can be computed in time $O(n \log k)$ since, for $j \in [k]$ we have, for some $i \in [n]$, that $\Pi B \Pi^T = R H (B (H e_i))$. Now, this computation involves (in sequence) a WHT applied to a 1-sparse vector in time $O(n)$ as per \cref{thm:1-sparse}, a scaling by $B$, and a trimmed WHT in $O(n \log k)$. Hence, the entire matrix can be computed in time $O(n k \log k)$.

Finally, the linear system can be solved in $O(k^3)$, yielding a total complexity of $O(n k \log k + k^3)$.
\end{proof}

\subsection{Matrix-vector multiplication}
\label{appendix:matrix-vector}

We do the computation in the order indicated in the final line below.
\begin{align}
A \beta
&= (\Pi^T K \Pi + \Pperp D \Pperp) \beta \nonumber \\
&= \left(\frac{n}{k} \Sigma H R^T K R H \Sigma + (I - \Sigma H R^T R H \Sigma) D (I - \Sigma H R^T R H \Sigma) \right) \beta \nonumber \\
&= D \beta - D \Sigma H R^T R H \Sigma \beta + \Sigma H R^T (Q R H \Sigma \beta - R H \Sigma D \beta) \nonumber \\
&= D \beta - D \Sigma H (R^T (\tPi \beta)) + \Sigma H (R^T (Q (\tPi \beta) - \tPi (D \beta))) \label{eqn:Abeta} .
\end{align}
For a vector $x$, $\tPi x$ can be computed in $O(n \log k)$ using the SRHT. Diagonal scaling (by $D$ and $\Sigma$) costs $O(n)$. Applying $Q$ to some $z \in \real^k$ costs $O(k^2)$. Finally, for $z \in \real^k$, observe that $H R^T z$ is the Walsh-Hadamard transform of a $k$-sparse vector (since $R^T z$ scatters the $k$ entries of $z$ into a $k$-sparse $n$-dimensional vector), which can be computed in time $O(n \log k')$. 

\subsection{Pseudo-code and Complexity of product of Walsh-Hadamard matrix with one-sparse vector}
\label{appendix:1-sparse-proof}

The following algorithms calculates $ r = H v $, where $ H \in \real^{n \times n}$ is a 
Walsh-Hadamard matrix, $ v $ a $1$-sparse vector, and $ n = 2^m $ because all Walsh-Hadamard matrices
have dimensions which are powers of $2$.
Assume the indices of $ v $ and $ H $ are zero based and that $ i $ is the index of the 
only non-zero component of $ v $. The algorithm utilizes the fact that the result is equal to
the $i$-th column of $ H $, denoted $h $, multiplied by the scalar $ v(i) $. 

The column $h $ of $ H $ can be found by using the recursive structure of the 
Walsh-Hadamard matrix
\begin{align} \label{eqn:recursiveH}
  H_{n} = \begin{bmatrix}
           H_{n/2} &   H_{n/2} \\
           H_{n/2} & - H_{n/2}
           \end{bmatrix}.
\end{align}
If the index $i$ is in the interval $ 0 <= i < \frac{n}{2} $ then the column $h $ will sit in the left part of
the block matrix in \eqref{eqn:recursiveH} and therefore the sub-vector of the first $\frac{n}{2}$ components
of $ h$ will be equal to the sub-vector of the second $\frac{n}{2}$ components,
$ h (\frac{n}{2}, \dots, n-1) = h (0, \dots,  \frac{n}{2}-1) $. 
If however $i$ is in the other interval $ \frac{n}{2} <= i < n $ then the column $h$ will be in the right
part of the block matrix of \eqref{eqn:recursiveH} and therefore the second sub-vector is the negative of the first
sub-vector,  
$ h (\frac{n}{2}, \dots, n-1) = - h (0, \dots,  \frac{n}{2}-1)$. Testing which of the two
cases occurs can be efficiently done by checking the value of bit $m$ of index $i$, where
we use the convention that the least-significant bit of $i$ is bit 0.

The problem has now been reduced to calculating the components $ h (0, \dots,  \frac{n}{2}-1) $
in the matrix $ H_{n/2} $ which can be done similarly by looking at the value of bit  $m-1$ of 
index $i$. The base case is reached when we set $ h (0) = 1 $ and then update according to
\begin{align}
  h (1) = \begin{cases}
           + h (0) & \text{if bit $0$ of $i$ is $0$} \\
           - h (0) & \text{if bit $0$ of $i$ is $1$}
           \end{cases}.
\end{align}
The result $r$ can therefore be constructed by initializing the first component of $r$ to $v(i)$ and
then copying $ 1 + 2 + \dots + \frac{n}{2} = n - 1 $ components of sub-vectors of the result $r$ into 
yet uninitialized sub-vectors of $r$ using the bits of the index $i$ to choose the appropriate sign. 
The complexity is therefore $O(n)$.

\begin{algorithm}[H]
  \DontPrintSemicolon
  \SetAlgoLined
  \KwIn{\begin{tabular}{|l l} 
            $n$                       & the dimension of the vector $ v$ \\ 
            $i \in \{0,\dots, n-1 \}$ & index of single non-zero component of $v$ \\ 
            $v(i)$                    & value of non-zero component of $v$ 
            \end{tabular}}
  \KwOut{$ r(0, \dots, n-1) $ where $ r = H v $}
  $ r(0) \leftarrow  v(i) $ \;
  $ m    \leftarrow  1 $ \;
  \While{$ m < n $}{
    \eIf{ $i \mod 2 = 0$}{
       $ r (m, \dots, 2*m - 1) \leftarrow \; + r (0, \dots, m-1)$\
      }{
       $ r (m, \dots, 2*m - 1) \leftarrow \; - r (0, \dots, m-1)$\
      }
     $ i \leftarrow i/2  $\;
     $ m \leftarrow m *2 $\;
    }
  \caption{Product of Walsh-Hadamard matrix with $1$-sparse vector}
\end{algorithm}

\end{document}